\documentclass{article}

\usepackage{PRIMEarxiv}

\usepackage[utf8]{inputenc} 
\usepackage[T1]{fontenc}    
\usepackage{hyperref}       
\usepackage{url}            
\usepackage{booktabs}       
\usepackage{amsfonts}       
\usepackage{nicefrac}       
\usepackage{microtype}      
\usepackage{lipsum}
\usepackage{fancyhdr}       
\usepackage{graphicx}       
\graphicspath{{media/}}     
\usepackage[ruled,longend]{algorithm2e} 
\usepackage{amssymb}
\usepackage{amsmath}
\usepackage{subcaption}
\usepackage{bm}

\SetKwComment{Comment}{ $\rhd$ }{ }

\title{Improved Batching Strategy For Irregular Time-Series ODE}

\author{
  Ting Fung Lam, Yony Bresler, Ahmed Khorshid, Nathan Perlmutter\thanks{former member} \\
  Crater Labs \\
  Toronto, Canada\\
  \texttt{\{wilton, ahmed, yony\}@craterlabs.io} \\
}

\begin{document}
\maketitle

\begin{abstract}
Irregular time series data are prevalent in the real world and are challenging to model with a simple recurrent neural network (RNN). Hence, a model that combines the use of ordinary differential equations (ODE) and RNN was proposed (ODE-RNN) to  model irregular time series with higher accuracy, but it suffers from high computational costs. In this paper, we propose an improvement in the runtime on ODE-RNNs by using a different efficient batching strategy. Our experiments show that the new models reduce the runtime of ODE-RNN significantly ranging from 2 times up to 49 times depending on the irregularity of the data while maintaining comparable accuracy. Hence, our model can scale favorably for modeling larger irregular data sets.
\end{abstract}


\section{Introduction}
Time series models are ubiquitous in numerous applications such as predictive maintenance \cite{Kanawaday2017, Lin2019Predictive, Kiangala2020}, financial forecasting \cite{masini2021machine}, and next sentence prediction \cite{mikolov2013efficient, devlin2018bert}. While statistical time series methods such as ARIMA can outperform machine learning methods for univariate datasets \cite{Makridakis2018}, machine learning promises superior performance for large multivariate datasets \cite{cerqueira2019machine}. Recurrent Neural Networks (RNN) with memory retention such as Gated Recurrent Units (GRU) or Long Short Term Memory (LSTM) have been the gold standard for temporal machine learning models. The hidden state $h_i$ of a traditional RNN cell is only updated every observation, which is not a problem if the data points are equidistant in space or time. However, this poses limitations for applications where observations are irregular or cannot be obtained periodically \cite{saeed2011multiparameter, johnson2016mimic}. For instance, various measurements of patients are often obtained at irregular intervals \cite{Silva2012Physionet}, so there is a need for a model that takes into account the time difference between data points. 

To account for irregular sampling, data points are often binned into regular intervals using aggregation. This technique has two problems: 1. The sampling interval chosen has to be larger than the smallest interval between two consecutive data points. 2. Averaging data points destroys valuable information about the correlation between sampling intervals and latent variables \cite{lipton2016directly, che2018recurrent}. Another solution is to include the time delta as an additional feature to the RNN input $h_i = \text{RNNCell} ({h_{i - 1}}, \Delta t, x_i)$, but this will not interpolate the hidden state between observations \cite{NEURIPS2019_42a6845a}. Rubanova \textit{et al.} have developed an RNN model that evolves the latent hidden states using ordinary differential equations (ODE-RNN) \cite{NEURIPS2019_42a6845a}. First, the ODE is solved using a Neural ODE, a class of neural networks with parameters $\theta$, which defines the latent state as a solution to an ODE initial-value problem \cite{NEURIPS2018_69386f6b}. The latent state at time $t_i$ is defined as ${h'_i} = \text{ODESolve} (f_{\theta}, h_{i-1}, (t_{i-1}, t_i))$ where $f_{\theta} = dh'/dt$. Second, the hidden state is updated using a standard RNN: $h_i = \text{RNNCell} ({h'_i}, x_i)$ \cite{NEURIPS2019_42a6845a}. These two steps are repeated for subsequent time steps. An ODE can be solved numerically using an adaptive step approach such as the Runge-Kutta method, or a fixed step approach such as the Euler method. The ODE solver from Rubanova's work can be used with both methods \cite{chen2021eventfn}. However, their approach requires using the union of time values across a mini-batch (combined time). This can drastically slow down the training time, especially with the presence of multiple non-unique time values. 

In this paper, we propose using a fixed step approach coupled with a batching strategy that scales linearly with the sequence length. We demonstrate that our proposed method can achieve both better accuracy and speed in comparison to the combined time approach on public and synthetic datasets.

\section{Related Work}

\paragraph{Problem statement}
We consider the auto-regressive problem: Given a time series $i \in \left\{0,...,\text{N-1}\right\}$ of input features $\mathbf{x}_i \in \mathbb{R}^{d_x}$ at times $t_i  \in \mathbb{R}$, given a time $t_{N}$, predict $\bm{y}_{N} \in \mathbb{R}^{d_y}$. We perform prediction at a single future point for simplicity. We denote in bold vectors or tensors that vary across the mini-batch, to distinguish them from scalars.

\paragraph{Combined time ODE-RNN model}
We begin by examining how Rubanova’s \cite{NEURIPS2019_42a6845a} combined method works to demonstrate how more irregular datasets lead to increasingly slower model training.  The reason for the slowdown is two-fold: The first is that while an RNN requires only a single forward step between each input sample, the ODE portion takes multiple steps that depend on both the size of the time jump $\Delta t_{N+1} = t_{N+1}-t_N$ as well as the curvature ($df_{\theta}/dt$) of the hidden state. The tuning of hyperparameters allows for some trade-off between training time and accuracy, by changing either the acceptable error tolerance (for adaptive procedures such as dopri5) or by changing the fixed time jump (for fixed procedures such as forward Euler). A second reason for slower training that is more difficult to sidestep by tuning hyperparameters is due to the way the combined time method handles integration time in a mini-batch. If the mini-batch only has a single sequence or if the irregular sequence is identical across the mini-batch, the process is straightforward, as the model updates the hidden state using the RNN of sample $x_i$ and then uses ODE evolver to advance from time $t_i$ to $t_{i+1}$. 

\begin{figure}[htp]
    \centering
    \includegraphics[scale=0.6]{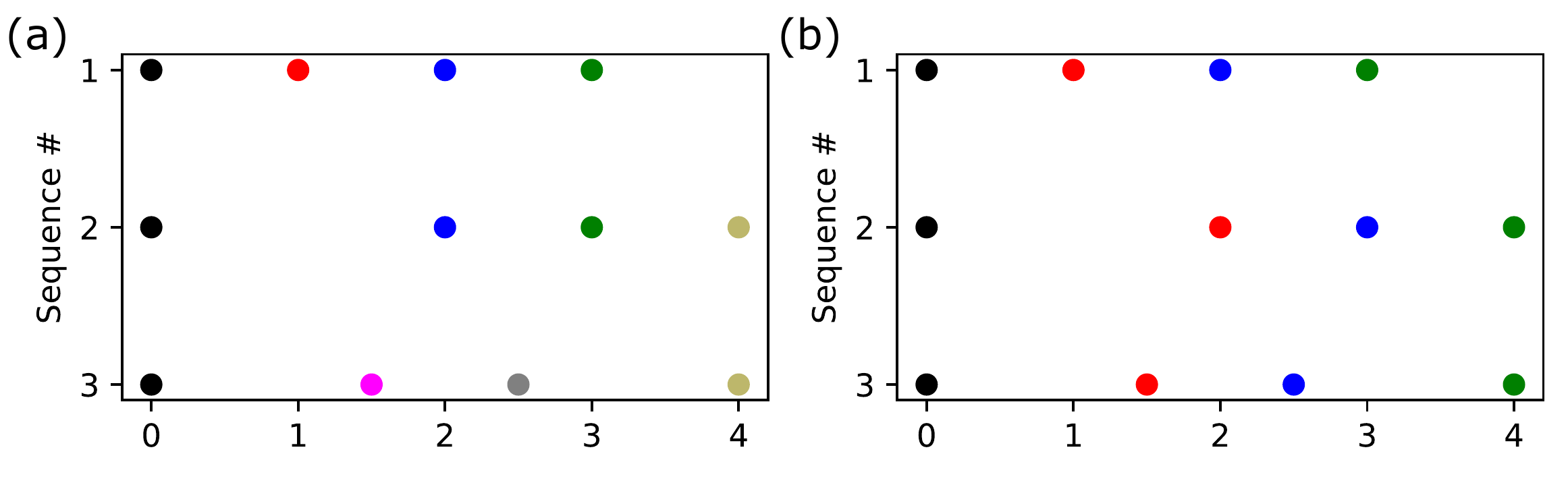}
    \caption{(a) The combined time method takes the union of all the time steps in a batch and loops every step. (b) Our model only loops every step regardless of the time values of the time steps.}
    \label{fig:evolver_loop}
\end{figure}

The process requires additional steps when the mini-batch has multiple samples that are not equal in their irregular intervals. Consider the illustration shown in Figure \ref{fig:evolver_loop}(a) for a mini-batch of 4, each with 4 samples at different times as indicated by the horizontal axis. Time is evolved in unison across the mini-batch. The sequence begins with all samples at $t_0 = 0$ (black), and the RNN with $x_0$ (in black) can be applied. Then the hidden state is evolved forward in unison across the mini-batch, as all 4 samples are integrated up to $t = 1$. An RNN update is performed only for the second sequence (shown in red), while the other two sequences are unchanged, using a mask. Then integration continues to $t = 1.5$, RNN is applied selectively to the third sequence (shown in pink), and the process continues until the last time point is reached. Pseudocode for this is given in Algorithm \ref{alg:odeintRNN}. This process slows training: the loop iterated over the unique times across the mini-batch which can be much larger than the length of any single item in the mini-batch. Additionally, masking and indexing are required at each iteration to track the RNN input features. This slowdown is more pronounced when using a large mini-batch size, or when the dataset is more irregular, and the number of unique times across the batch grows. 

\begin{algorithm}
\KwData{Data points $\{\bm{x}_i\}_{i=0..N-1}$, corresponding time differences $\{\bm{\Delta t}_i\}_{i=0..N-1}$ and final jump $\{\bm{\Delta t}_i\}_{i=N}$}
$\{\bm{t_i}\}_{i=0..N} = {\sum_{j=0}^{i}{\bm{\Delta t_j}}}$\;
$ct = \text{unique}(\bm{t_n})$ \Comment*[r]{combined
time across mini-batch}
$\bm{h} = 0$ \;
\For{$j = 0..\emph{len(ct)}-1$}{
    $t_{cur} = ct\left(j\right)$ \;
    $t_{prev} = ct\left(j-1\right)$ \;
    $\bm{h}' = 
        \begin{cases} 
            \text{ODESolve}(f_{\theta}, \bm{h}, (t_{prev}, t_{cur})), & t_{cur} < \bm{t}_{N+1} \\
            \bm{h}, & otherwise
        \end{cases} 
    $ \Comment*[r]{Perform ODE if needed}
    
    $ \bm{h} = 
            \begin{cases} 
            \text{RNN}(\bm{h}', \bm{x}(\bm{t} = t_{cur})), & t_{cur} \in \bm{t} \\
            \bm{h}', & otherwise
        \end{cases}
    $ \Comment*[r]{Update hidden state if it exists for $t_{cur}$}
}
\KwResult{$\bm{h}$}
\caption{Rubanova's combined time ODE-RNN}
\label{alg:odeintRNN}
\end{algorithm}
\paragraph{Other related work}
Recent works used various strategies to improve the efficiency and accuracy of ODEs for learning irregular time series data. The second-order neural ODE optimizer (SNOpt) computes second-order derivative gradients for the backward propagation to improve training efficiency \cite{NEURIPS2021_d4c2e4a3}. Likewise, heavy ball ODE (HBODE) combines gradient descent with momentum with ODE, to generate a 2nd order ODE with a damping factor to accelerate training \cite{NEURIPS2021_9a86d531}. The Taylor-Lagrange Neural ODE (TL-NODE) model uses the Taylor expansion series with Lagrange form of error approximation of the Taylor series to replace the ODESolve for doing numerical integration, to improve the efficiency of the forward propagation of neural ODE \cite{djeumou2022taylor}. The model order reduction method uses proper orthogonal decomposition to reduce the dimensions of the weight matrices of the network, and a discrete empirical interpolation Method to replace the activation functions with interpolation operations \cite{lehtimaki2021accelerating}. Lyanet uses the Lyapunov theory from control theory to replace the loss function of neural ODE with the Lyapunov Loss to guarantee that the ODE exponentially converse and provide adversarial robustness \cite{rodriguez2022lyanet}. Skip DEQ made two improvements to the existing DEQ model by first, add an explicit layer with some regularizations before the implicit layer to better predict the initial state, and secondly, replace the implicit layer with an ODE that runs to infinite time using adaptive ODE solvers \cite{pal2022mixing}. Finally, the neural flows method does not approximate the solution of an ODE but directly learns the solutions of the ODE, so an ODE solver is not necessary \cite{bilos2021neuralflows}.

\section{Model}

Our approach is to reduce training time performance by changing how time is evolved across the batch. Instead of using the combined unique times, we allow each sample in the mini-batch to evolve independently, as shown in Figure \ref{fig:evolver_loop}(b). As before, all sequences begin at $\bm{t_0}=0$, and the RNN $(\bm{x}_0)$ is applied. But now each sequence across the mini-batch is evolved to different points in time: $\bm{t}_1 = \left[1, 2, 1.5\right]$. The RNN for the observations $\bm{x}_1$  (shown in red) can then be applied to all sequences in the batch, each sample is evolved to its next sample in time (shown in blue), and the process continues. Full details are shown in Algorithm \ref{alg:odernn}. This can have several advantages: First, the number of iterations of the loop is set by the single longest input sequence, and not the combined unique time points across the entire mini-batch. Secondly, it eliminates the overhead of masking the RNN updates, since they occur simultaneously across the mini-batch.

\begin{algorithm}[ht]
\caption{Our batch efficient ODE-RNN}
\label{alg:odernn}
\KwData{Data points $\{\bm{x}_i\}_{i=0..N-1}$, time differences $\{\bm{\Delta t}_i\}_{i=0..N-1}$, and final time-jump $\{\bm{\Delta t}_i\}_{i=N}$}
$\bm{h} = \textbf{0}$\;
\For{$j = 0..N-1$}
{
    $\bm{h}' = $ Evolver$(\bm{h}, \bm{\Delta t}_j$) \Comment*[r]{Refer to Evolver Algorithm \ref{alg:evolver}/\ref{alg:evolver_af}/\ref{alg:evolver_geometric}}
    $\bm{h} = $ RNN$(\bm{x}_j, \bm{h}')$\;
}
$\bm{h} = $ Evolver$(\bm{h}, \bm{\Delta t}_N$) \Comment*[r]{Perform final-jump before making prediction}
$\bm{o} = \text{OutputNN}(\bm{h})$\;
\KwResult{$\bm{o}, \bm{h}$}
\end{algorithm}

The combined time approach with the Torchdiffeq evolver allows for many choices in the integration method: fixed time-step methods such as forward Euler, the midpoint method or Runge-Kutta, and adaptive integrators where a relative error rate is given and the method automatically adjusts the time step to meet the estimated threshold. Keeping our objective of improving model performance, we introduce three different evolver modules, all of which use a form of forward Euler, each with a slightly different method of performing the integration. The first uses a fixed time step, with a varying number of steps across the mini-batch. The second uses a fixed number of steps, with varying time steps across the mini-batch. The third uses a geometrically increasing time step, with a varying number of steps across the mini-batch. Our approach is best suited for large irregular time-series datasets where the irregularity varies across mini-batch samples. 

Our first model is the fixed dt method that uses a scalar value $\Delta t$, which is held constant throughout training, to evolve all hidden states.  As shown in Algorithm \ref{alg:evolver}, the loop is set by the largest time to evolve in the batch, and a simple mask is used to prevent updating any hidden state once its respective $\Delta t_i$ is reached. This ensures the rounding error is independent of the time jumps but can lead to long run-time and very deep networks when the largest time jump is much larger than $\Delta t$. 

\begin{algorithm}[ht]
\caption{Evolver module: Fixed dt mode}
\label{alg:evolver}
\KwData{Step size $s$, Hidden state $\bm{h}$ and corresponding time differences $\bm{\Delta t}$}
$\bm{n} = \bm{\Delta t} / s$ \Comment*[r]{number of steps varies across mini-batch}
$N = max(\bm{n})$\;
\For{$j = 0..N-1$}
{
    $\textbf{mask} = $ j $ < \bm{n}$\;
    $\bm{h}' = f_{\theta}(\bm{h})$\Comment*[r]{learning $\frac{d\bm{h}}{dt}$}
    $\bm{h} = \bm{h} + s * \textbf{mask} * \bm{h}'$
}
\KwResult{$h$}
\end{algorithm}

\begin{algorithm}[ht]
\caption{Evolver module: adaptive fixed mode}
\label{alg:evolver_af}
\KwData{Number of steps $N$, Hidden state $\bm{h}$ and corresponding time differences $\bm{\Delta t}$}
$\bm{t_i} = {\sum_{j=0}^{i}{\bm{\Delta t_j}}}$\;
$\bm{s} = \bm{\Delta t} / N$ \Comment*[r]{step size varies across mini-batch}
\For{$ j = 0..N-1$}
{
    $\textbf{mask} = j * \bm{s} < \bm{\Delta t}$\;
    $\bm{h}' = f_{\theta}(\bm{h})$\Comment*[r]{FC learning $\frac{d\bm{h}}{dt}$}
    $\bm{h} = \bm{h} + \bm{s} * \textbf{mask} * \bm{h}'$
}
\KwResult{$h$}
\end{algorithm}

A second method uses a fixed number of iterations in the loop that is independent of the time differences in the mini-batch, shown in Algorithm \ref{alg:evolver_af}. Here the step size $s$ is no longer a scalar but a vector with values for each sequence in the mini-batch. This has the advantage of giving a consistent run-time and depth to the forward pass regardless of the time-jumps, though it can lead to larger rounding errors for larger valued $s$. 

The fixed dt and adaptive fixed methods have opposing trade-offs: the first can suffer from very deep networks, while the second may incur large rounding errors.
We propose a third method in an attempt to balance the two. We hypothesize that there is more change in curvature in the hidden state immediately after a new measurement $t ~= t_i$, than there is if the state is evolved for a later time $t >> t_i$. The geometric adaptive algorithm starts with a small step size $s_0$, which is then increased using the multiplicative constant $s_i = s_{i-1} * r$ with the growth factor $r>1$. If the time step were to overshoot $t_{i+1}$, it is reduced to reach it exactly, and a mask is used to prevent it from being evolved further. Full details are shown in Algorithm \ref{alg:evolver_geometric}. For example, using our default values $s_0 = 0.001, r=1.5$, requires $5$ steps to reach $t = 0.01$, $15$ to reach $t=1$, and only $39$ steps to reach $t = 10,000$. Although a similar increase in step size for a flatter derivative is possible with the adaptive routines of the combined time method, in practice
 we found that method to be even slower than using the combined time method with a fixed step size.

\begin{algorithm}
\caption{Evolver module: adaptive geometric mode}
\label{alg:evolver_geometric}
\KwData{Initial step size $s_0$, Growth factor $r > 1$, Hidden state $\bm{h}$ and corresponding time differences $\bm{\Delta t}$}
$N = \max \left\{ \lceil \log \left[ \left(r-1\right) * \bm{\Delta t} / s_0 \right] / s_0 \rceil \right\}$ \Comment*[r]{Find geometric series steps needed for largest value}
$t_{\text{cur}} = 0$\;
$s = s_0$\;
\For{$ j = 0..N-1$}
{
    $\textbf{mask} = j * s < \bm{\Delta t}$\;
    $\bm{s}_{\text{cur}} = \min\left\{s, \bm{\Delta t} - t_{\text{cur}} \right\}$   \Comment*[r]{Prevent overshooting target time if $s$ is too large}
    $\bm{h}' = f_{\theta}(\bm{h})$\Comment*[r]{FC learning $\frac{d\bm{h}}{dt}$}
    $\bm{h} = \bm{h} + \bm{s}_{\text{cur}} * \textbf{mask} * \bm{h}'$\;
    $t_{\text{cur}} \mathrel{+}= s$\;
    $s = s * r$\Comment*[r]{Grow $s$ by factor $r$}
}
\KwResult{$h$}
\end{algorithm}

\section{Experiments}
We evaluate our model on synthetic and real-world datasets including synthetic sine waves, MuJoCo physics simulation, and MIMIC-IV clinical dataset. The task for all of our models is to predict the last time step given all the preceding time steps in the dataset. They are trained to minimize the MSE loss in accordance with the hyperparameters and details listed in the appendix. In the following section, we compare the training time and accuracy of these three models to a simple RNN model as well as the combined time method baseline, using several datasets. We implement the combined time method with a wrapper for torchdiffeq \cite{NEURIPS2018_69386f6b} \footnote{The code will be available online publicly soon.}.

\subsection{Synthetic dataset}
\begin{figure}[htp]
    \centering
    \includegraphics[scale=0.5]{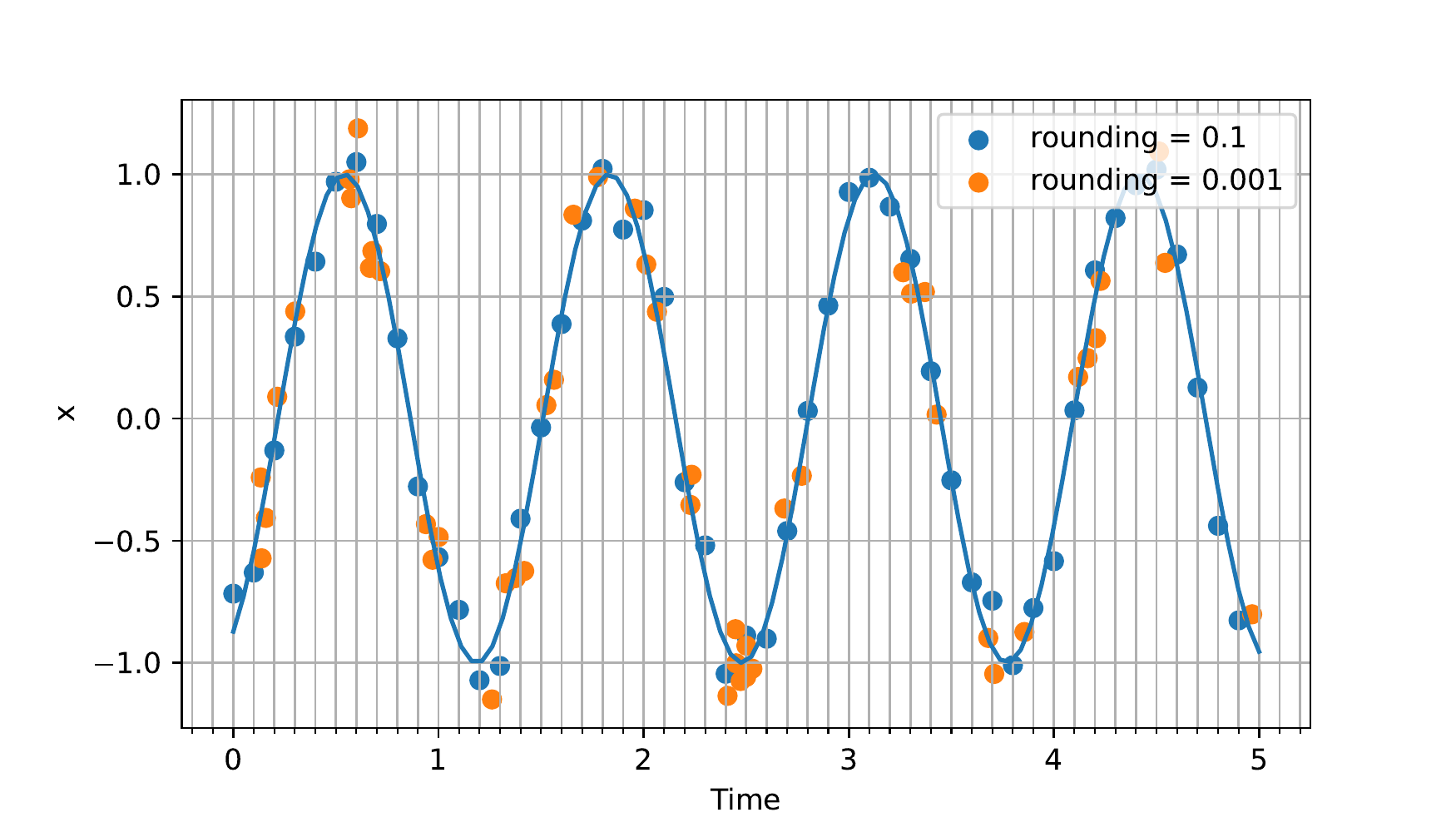}
    \caption{Points sampled from a sample sine function trajectory with step rounding of 0.1 and 0.001 time units for comparison, with vertical grid lines of interval of 0.1 units.}
    \label{fig:syn_plot}
\end{figure}

We follow the procedure stated by Rubanova \textit{et al.} to generate 10,000 synthetic sinusoidal wave sequence data  with variable frequencies and starting position, and constant amplitude with a few minor modifications \cite{NEURIPS2019_42a6845a} (see Appendix \ref{sec:syn}). To find out how the irregularity of the data influences the training time and accuracy, we generate random sequences with 50 random time steps in each sequence. The time steps are sampled with rounding of 0.1 and 0.001 units for comparison (Figure \ref{fig:syn_plot}).

Our efficient ODE-RNN models achieve superior performance in terms of run time and accuracy against the baseline combined time method and simple RNN models. Regarding the different modes of our model, the adaptive fixed mode gives the least mean squared error (MSE) for time values rounded to the nearest 0.001 units, while all models result in similar MSEs for rounding of 0.1 units. Regarding the 0.001 units rounded dataset, our model achieves a significant accuracy improvement compared to the combined time method and simple RNN due to the high irregularity of the dataset. Moreover, the combined time method also attains significant improvement in accuracy over simple RNN. In addition, we observe that the dopri5 solver offers only minimal improvements in accuracy for the combined time method model as demonstrated in previous research \cite{bilos2021neuralflows}, while using greater tolerance parameters to make training feasible (Table \ref{tab:acc}). 

We achieve a speed-up of training time per epoch of at least two times and up to 49 times with our ODE-RNN model for the synthetic datasets. In addition, the time steps with rounding of 0.001 units take significantly longer for the combined time method model to train while slightly longer for our model. Since the number of unique time steps in a batch with a rounding of 0.001 units is much greater than for a rounding of 0.1 units, more iterations are required, thus resulting in a tremendous increase in training time per epoch. As a result, our method achieves a speedup of 49 times against the combined time method using the Euler solver. Also, using multi-GPU fails to speed up the training process due to the relatively small batch size being used. For the combined time method models, we use a larger learning rate of 0.01 as in Rubanova \textit{et al.} to reduce the number of epochs necessary, so that the training can be feasibly performed. \cite{NEURIPS2019_42a6845a} Using dopri5 solver for the combined time method model results in approximately 4 times longer training runtime than the Euler solver while achieving no significant improvements as demonstrated previously \cite{bilos2021neuralflows} (Table \ref{tab:runtime}). 

\begin{table}[htp]
\begin{tabular}{l|cc|cc|c}
                   & \multicolumn{2}{l|}{Synthetic (Rounding)} & \multicolumn{2}{l|}{MuJoCo (\% points   selected,  $\times 10^{-3}$)} & MIMIC-IV          \\
MSE                & 0.001                & 0.1                 & 50\%                              & 10\%                              &                  \\ \hline
Combined time (euler)     & 0.05722              & 0.01223             & 0.30516                           & 1.94714                           & n.d.             \\
Combined time (dopri5)    & 0.04780              & \textbf{0.01218}    & 0.28009                           & 1.94433                           & n.d.             \\
Simple RNN           & 0.10344              & 0.01264             & 0.21902                           & 1.96752                           & 0.27685          \\ \hline
Fixed dt            & 0.02832              & 0.01261             & 0.22483                           & 1.26604                           & \textbf{0.27527} \\
Adaptive fixed     & \textbf{0.01295}     & 0.01249             & 0.25461                           & \textbf{1.25475}                  & 0.27648          \\
Adaptive geometric & 0.01367              & 0.01261             & \textbf{0.21853}                  & 1.37623                           & 0.27681         
\end{tabular}
\caption{Test mean squared error for all datasets. Best results are bolded. All runs are performed with fixed random seeds to make the results reproducible.}
\label{tab:acc}
\end{table}

\begin{table}[htp]
\begin{tabular}{l|rr|rr|r}
                        & \multicolumn{2}{c|}{Synthetic (Rounding)} & \multicolumn{2}{c|}{MuJoCo (\% points selected)} & MIMIC-IV          \\
Wall time per epoch (s) & 0.001                 & 0.1                & 50\%                    & 10\%                   &                  \\ \hline
Combined time (euler)          & 1164.80$\pm$34.88     & 32.26$\pm$1.18     & 60.27$\pm$2.40          & 59.86$\pm$2.10         & n.d.             \\
Combined time (dopri5)         & 4414.27$\pm$297.87    & 137.10$\pm$6.72    & 204.15$\pm$9.01         & 215.34$\pm$7.51        & n.d.             \\
Simple RNN                & 11.36$\pm$0.26        & 8.54$\pm$0.24      & 9.75$\pm$0.43           & 3.54$\pm$0.16          & 113.21$\pm$1.52  \\ \hline
Fixed dt                 & 24.94$\pm$1.06        & 15.37$\pm$0.64     & 22.62$\pm$0.94          & 18.03$\pm$0.75         & 144.89$\pm$2.23  \\
Adaptive fixed          & 29.58$\pm$1.23        & 27.45$\pm$0.36     & 29.12$\pm$1.28          & 8.84$\pm$1.18          & 499.12$\pm$19.91 \\
Adaptive geometric      & 23.72$\pm$1.42        & 16.75$\pm$0.59     & 24.23$\pm$0.81          & 11.66$\pm$0.21         & 195.09$\pm$2.32 
\end{tabular}
\caption{Average training runtime per epoch after the first epoch for all datasets with standard deviation.}
\label{tab:runtime}
\end{table}

\begin{figure}[htp]
    \centering
    \includegraphics[scale=0.5]{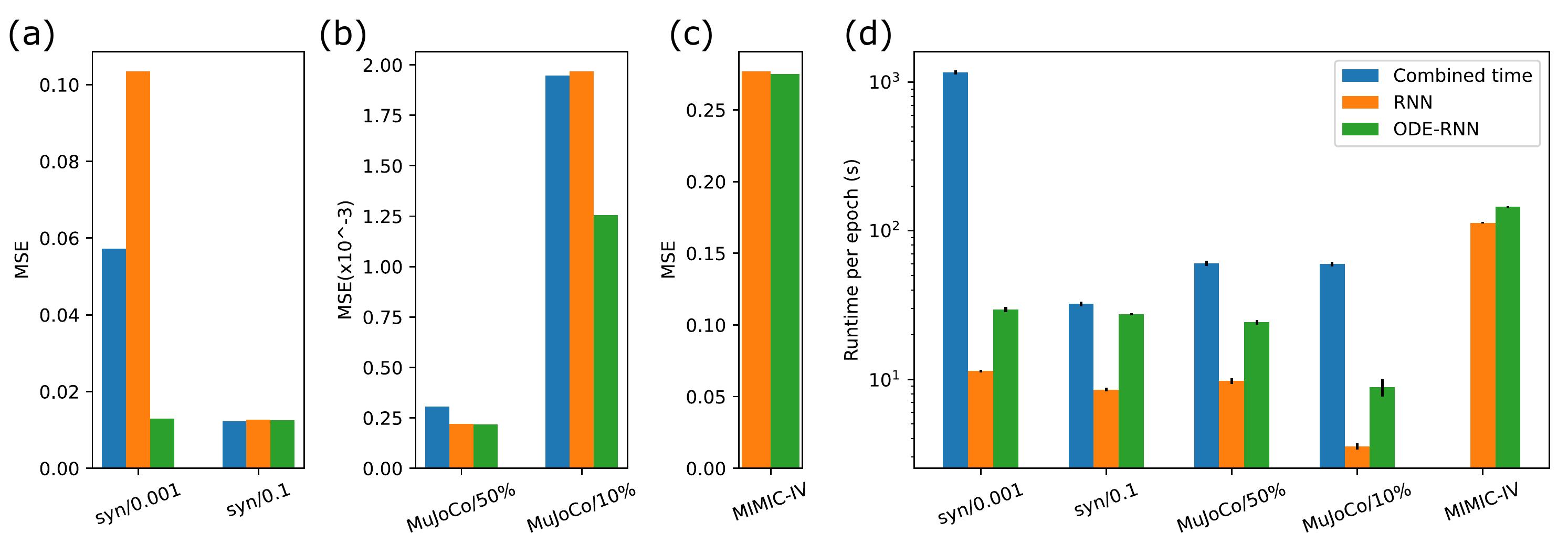}
    \caption{Test MSE of a) synthetic dataset with rounding b) MuJoCo dataset with percentage of points selected c) MIMIC-IV dataset. d) Average runtime per epoch of all datasets with standard deviation. For ODE-RNN, the models which result in the lowest MSE are plotted}
    \label{fig:plot}
\end{figure}

\subsection{MuJoCo physics simulation}
MuJoCo is a physics engine containing a module named "Hopper" for generating the physical dynamics of an imaginary frog-like organism. Following the previously reported methods exactly \cite{NEURIPS2019_42a6845a, bilos2021neuralflows}, 10,000 sequences of the "hopper" with 100 time steps and 14 features in each step are generated. Subsequently, we randomly sample 50\% and 10\% of the time steps for performing comparison experiments and use them to train our model.

Our model attains better performance in terms of accuracy against the baseline and simple RNN models in a similar fashion to the synthetic datasets. Moreover, the different methods of our model perform similarly, and the model achieves maximum performance gain for the 10\% sampled dataset, which has higher irregularity than the 50\% sampled dataset (Table \ref{tab:acc}). In addition, our implementation results in at least two times faster training runtime than the combined time method (Table \ref{tab:runtime}).

\subsection{MIMIC-IV clinical dataset}
The MIMIC-IV dataset contains clinical data of more than 40,000 intensive care unit patients in Beth Israel Deaconess Medical Center \cite{mimiciv_v1}. We follow the procedure stated in \cite{bilos2021neuralflows} without any changes to process the MIMIC-IV v1.0 dataset so that it contains time series data of 17,874 patients and 102 features. 

Our results indicate that all methods of our model attain only an insignificant improvement in terms of MSE compared to simple RNN (Table \ref{tab:acc}). Since we observe that our model overfits the training data in higher degrees compared to RNN, we attempt to address it by adding dropout layers to our ODE evolver to reduce overfitting. Nonetheless, our model is unable to achieve any significant improvements compared to simple RNN. A possible reason is that the observed features of the MIMIC-IV data are too sparse for the models to train in contrast to the synthetic and MuJoCo datasets. Notably, the results in \cite{NEURIPS2019_455cb265} also show that the neural ODE-VAE models perform worse than simple RNN for the similar MIMIC-III dataset. Meanwhile, the combined time model has a much higher runtime per epoch, and the training fails to converge for this dataset.

\section{Discussion and conclusion}

We propose the new ODE-RNN models to improve the combined time method, including fixed dt, adaptive fixed, and adaptive geometric modes. The new mini-batching strategy allows faster computation by reducing the number of iterations necessary during training. Hence, these models improve the speed of ODE-RNN without sacrificing its accuracy, as demonstrated by our experiments on both synthetic and real-life data. We believe that this method works because unlike solving a regular ODE, the model learns $f_\theta$ and may be able to compensate for coarser time steps. Our models can be scaled for processing larger irregular time-series datasets at a more reasonable financial cost. Finally, our models are more likely to achieve higher performance gain compared to simple RNN when the irregularity of the input dataset increases, so it should be utilized on datasets with high irregularities.
 
In this paper, our task is to predict the last data point from a time series sequence. Therefore, our models can be extended to different tasks such as interpolation and extrapolation for more than one data point. This will facilitate a wider range of real-life applications. It will be interesting to test our model on large and irregular datasets that are not sparse, as well as extending the evolver to use higher-order methods.

We test our model on non-public datasets that are highly irregular and common in the industry. In one experiment, the dataset consists of phone call logs with features such as call duration, call direction, call type, and time of the call. The task is to predict the duration of the call after a sliding window of calls. Due to the high cardinality of unique times within the call logs mini-batch, our model is able to achieve similar accuracy to the combined time approach at a fraction of the processing cost.

\section*{Acknowledgments}
We thank Seyed-Parsa Hojjat for useful discussions and feedback, and Nausheen Fatma for contributing to the combined times wrapper code. We also thank David Duvenaud and Yulia Rubanova for the helpful discussion at the onset of our project.

\bibliographystyle{ieeetr}  
\bibliography{references}  

\clearpage

\renewcommand{\thefigure}{A\arabic{figure}}
\setcounter{figure}{0}
\renewcommand{\thetable}{A\arabic{table}}
\setcounter{table}{0}

\appendix

\section{Data pre-processing}

\subsection{Synthetic data}
\label{sec:syn}
The protocol reported by Rubanova \textit{et al.} is followed \cite{NEURIPS2019_42a6845a}. We generate 10,000 one-dimensional sequences with 50 random time points rounded to the nearest of 0.1 as well as 0.001 units in the interval [0, 5]. We use the sine function with amplitude of 1 and frequency sampled uniformly from the interval of [0.5, 1], and sample the starting point from a normal distribution with a mean of 1 and standard deviation of 0.1.

\subsection{MuJuCo physics simulation}
The code published by Bilos \textit{et al.} and Rubanova \textit{et al.} is used directly to obtain the data for this experiment \cite{bilos2021neuralflows, NEURIPS2019_42a6845a}. They generate 10,000 sequences of the "hopper" with 100 time steps with 14 features. We use that data without any changes, and sampled 50 or 10 time steps randomly for the experiment.

\subsection{MIMIC-IV clinical dataset}
We followed \cite{bilos2021neuralflows} for processing MIMIC-IV dataset without any changes. They also follow and modify the procedure from \cite{NEURIPS2019_455cb265} which pre-processes MIMIC-III to pre-process MIMIC-IV. After running the code, the data contains time series of 17,874 patients with 102 features. We zero-padded the time series sequences to the longest sequence of 919 time steps at the beginning of the sequences. Also, the data was  used directly without any additional normalization. Because a lot of the features are not observed in each time step, the data contains a mask dataset of 102 features consisting of 0s and 1s indicating whether that corresponding features are observed or not.

\section{Training}

\subsection{General procedure}
All the data used in the experiments below are split into train, validation, and test sets with an 80:10:10 ratio. The models are trained with early stopping using the validation set, and the results are reported for the test set. We used a machine with 187 GB of RAM, as well as one or two NVIDIA Quadro RTX 8000 48GB GPUs to train our models for single and multi GPU modes respectively. We find the optimal hyperparameters for each model individually as time and available computation resources allow. We use a fixed random seed for all runs.
\subsection{Packages}
Please refer to \texttt{requirements.txt} in our code for the complete list of packages.
\begin{itemize}
    \item pytorch-lightning: 1.5.10
    \item pytorch: 1.10.2
    \item CUDA: 11.3
    \item torchdiffeq: 0.2.2 \cite{NEURIPS2018_69386f6b}
\end{itemize}

\section{Default hyperparameters for both our model and the combined time method}

\begin{itemize}
    \item All experiments: Adam optimizer
    \item Multi-GPU: False
    \item Early stopping patience epochs: 10
    \item Min/max epochs: 50/1000
    \item Learning Rate: 0.01 (Combined time) 0.001 (All other models)
    \item Default Mode: Fixed dt
    \item Number of adaptive steps for adaptive fixed mode: 5
    \item Dynamic step growth factor for adaptive geometric mode: 1.5
    \item Combined time method dopri5 tolerances: $10^{-3}$ (rtol) $10^{-4}$ (atol)
\end{itemize}

\subsection{Synthetic data}
\begin{itemize}
    \item Batch size: 50 (single-GPU), 25 (multi-GPU)
    \item Hidden size: 10
    \item Evolver dropout: False
    \item Step size: 0.1 (ODE-RNN fixed dt and Combined time with euler)
\end{itemize}

\subsection{MuJoCo physics simulation}
\begin{itemize}
    \item Batch size: 50 (single-GPU), 25 (multi-GPU)
    \item Hidden size: 20
    \item Evolver dropout: False
    \item Step size: 0.01 (ODE-RNN fixed dt and Combined time with euler)
\end{itemize}

\subsection{MIMIC-IV clinical dataset}
\begin{itemize}
    \item Batch size: 100
    \item Hidden size: 64
    \item Evolver dropout: 0.4
    \item Step size: 0.05 (ODE-RNN only)
\end{itemize}

\section{Additional results}
\begin{table}[!htp]
\begin{tabular}{l|rr|rr}
                              & \multicolumn{2}{c|}{Synthetic (Rounding)} & \multicolumn{2}{c}{MuJoCo (\% points selected)}  \\
Wall time per epoch (s)       & 0.001                & 0.1                 & 50\%                   & 10\%                   \\ \hline
Fixed dt, multi-GPU            & 26.13$\pm$1.19       & 16.55$\pm$0.40      & 22.76$\pm$0.48         & 18.32$\pm$0.45         \\
Adaptive fixed, multi-GPU     & 31.05$\pm$1.58       & 32.47$\pm$1.25      & 29.35$\pm$1.09         & 10.53$\pm$0.22         \\
Adaptive geometric, multi-GPU & 23.75$\pm$0.61       & 18.09$\pm$0.41      & 23.73$\pm$0.65         & 10.36$\pm$0.26        
\end{tabular}
\caption{Training runtime per epoch after the first epoch for multi-GPU mode using synthetic and MuJoCo datasets with standard deviation. Due to the large size of the dataset, our instrument does not have enough memory to run the multi-GPU modes.}
\end{table}

\end{document}